\documentclass[iicol,sn-mathphys-num]{sn-jnl}% Math and Physical Sciences Numbered Reference Style sn-mathphys-num, 
\usepackage{graphicx}%
\usepackage{multirow}%
\usepackage{amsmath,amssymb,amsfonts}%
\usepackage{amsthm}%
\usepackage{mathrsfs}%
\usepackage[title]{appendix}%
\usepackage{xcolor}%
\usepackage{textcomp}%
\usepackage{manyfoot}%
\usepackage{booktabs}%
\usepackage{ragged2e}
\usepackage{algorithm}%
\usepackage{algorithmicx}%
\usepackage{algpseudocode}%
\usepackage{listings}%
\usepackage{diagbox}%
\usepackage{tabularx}%
\usepackage{array}%
\usepackage{booktabs}%
\usepackage{adjustbox}%
\usepackage{microtype}%
\usepackage{makecell}%
\usepackage{hyperref}
\usepackage{xurl}
\usepackage{lmodern}
\begin{document}

\title[Article Title]{%Reflections Before Judgments: Enhancing User Assessment of Assistive Robots
Designing for Difference: How Human Characteristics Shape Perceptions of Collaborative Robots}%Assessing the perception of human-robot collaboration in dependence on human protected characteristics}

%%=============================================================%%
%% GivenName	-> \FYm{Joergen W.}
%% \author*[1,2]{\FYm{Joergen W.} \spfx{van der} \sur{Ploeg} 

\author[1,2]{{Sabrina} {Livanec}}
\equalcont{\small These authors contributed equally to this work.}

\author[1]{{Laura} {Londoño}}
\equalcont{\small These authors contributed equally to this work.}

\author[2]{{Michael} {Gorki}}

\author[1,3]{{Adrian} {Röfer}}

\author[1,3]{\\{Abhinav} {Valada}}

\author[2]{{Andrea} {Kiesel}}

\affil[1]{\orgdiv{BrainLinks-BrainTools Center}, \orgname{University of Freiburg}, \country{Germany}}
\affil[2]{\orgdiv{Department of Psychology}, \orgname{University of Freiburg}, \country{Germany}}
\affil[3]{\orgdiv{Department of Computer Science}, \orgname{University of Freiburg}, \country{Germany}}


\abstract{The development of assistive robots for social collaboration raises critical questions about responsible and inclusive design, especially when interacting with individuals from protected groups such as those with disabilities or advanced age. Currently, research is scarce on how participants assess varying robot behaviors in combination with diverse human needs, likely since participants have limited real-world experience with advanced domestic robots. In the current study, we aim to address this gap while using methods that enable participants to assess robot behavior, as well as methods that support meaningful reflection despite limited experience. In an online study, 112 participants (from both experimental and control groups) evaluated 7 videos from a total of 28 variations of human-robot collaboration types. The experimental group first completed a cognitive-affective mapping (CAM) exercise on human-robot collaboration before providing their ratings. Although CAM reflection did not significantly affect overall ratings, it led to more pronounced assessments for certain combinations of robot behavior and human condition. Most importantly, the type of human-robot collaboration influences the assessment. Antisocial robot behavior was consistently rated as the lowest, while collaboration with aged individuals elicited more sensitive evaluations. Scenarios involving object handovers were viewed more positively than those without them. These findings suggest that both human characteristics and interaction paradigms influence the perceived acceptability of collaborative robots, underscoring the importance of prosocial design. They also highlight the potential of reflective methods, such as CAM, to elicit nuanced feedback, supporting the development of user-centered and socially responsible robotic systems tailored to diverse populations.}




\maketitle

\section{Introduction}\label{sec1}

Robots have become essential in many contexts, including manufacturing, construction, healthcare, rehabilitation, and education. As advances in AI-based robotics increasingly enable robots to be used not only in highly controlled settings but also in human-populated environments, the demand for advanced human-robot interaction is becoming increasingly important. The integration of robots into human-centered work and leisure environments requires a balance between technical as well as human-centered functionality to ensure social acceptance~\cite{forlizzi2007robotic, hurtado2021learning}. 

\begin{figure*}
    \centering
    \includegraphics[width=\linewidth]{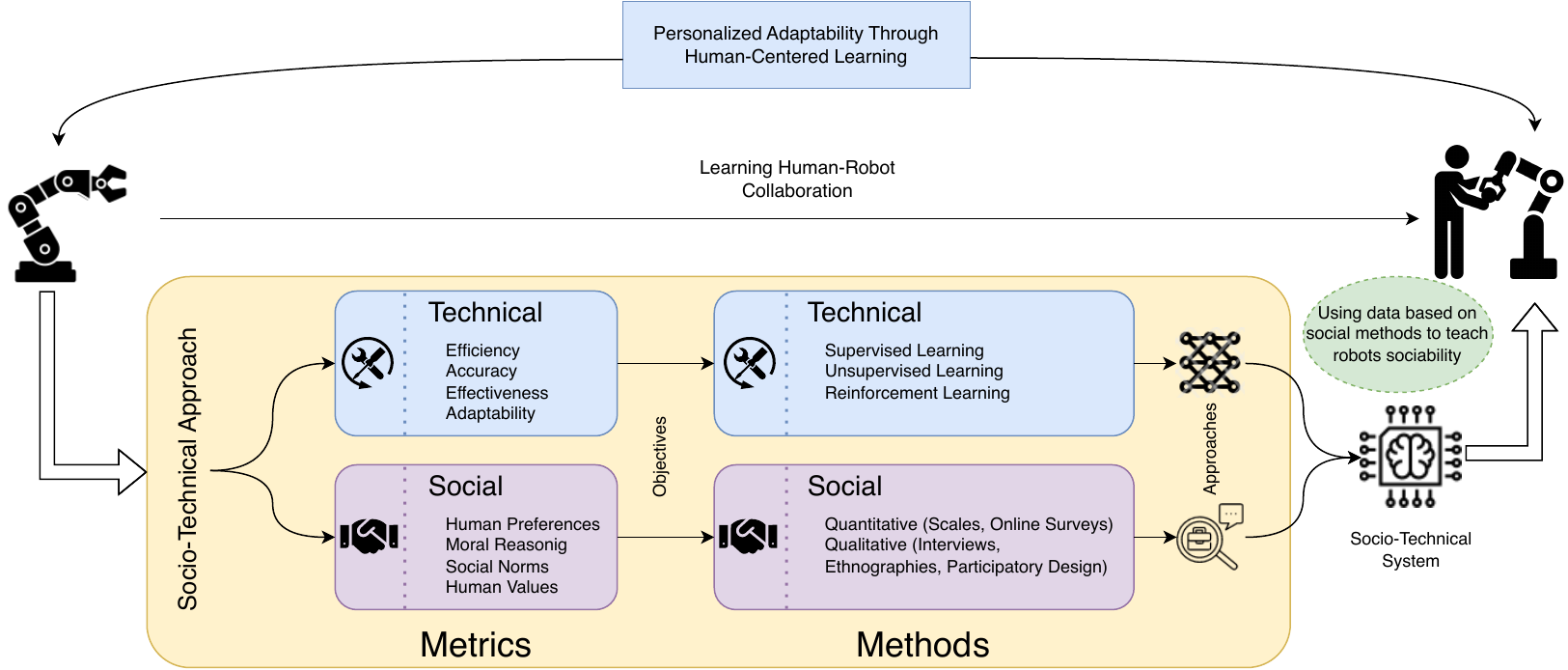}
    \caption{Human-Centered Design: A socio-technical approach for teaching sociability to robots, leveraging learning methods and data from social studies.}
    \label{fig:Figure 15.drawio.png}
\end{figure*}

Within the field of social robotics, human-robot collaboration (HRC) is emerging as a significant area of research with numerous applications in the service, social, manufacturing, and industrial context~\cite{coronado2022evaluating, Inga.2022}. However, since robots are expected to work alongside and collaborate directly with humans, major concerns arise regarding safety and ethics, as well as ensuring human comfort. Hence, robots are expected not only to operate efficiently and safely in human environments but also to adapt to the personal characteristics of the individuals with whom they are expected to collaborate. Establishing interactions between humans and robots that feel natural in the sense of human-like and socially appropriate is considered a crucial step towards achieving these goals~\cite {xie2023chatgpt}. 

As humans exhibit distinct preferences when engaging and collaborating with others, shaped by both cultural heritage and individual attributes, such as gender, age, or disabilities, etc., diverse groups hold varying perceptions and expectations of adaptive robots, influenced by their personal and cultural characteristics~\cite{correia2019choose, you2023trusting, mangin2022helpful}. Some studies suggest that, from the user's perspective, interaction with a robot is similar to interaction with fellow humans~\cite{kramer2012human, article, vanzoelen2021}. Given that collaborative robots are expected to operate in close proximity to and interact directly with individuals, it is essential to adopt a human-centered perspective in robot design to address the specific needs, preferences, and expectations humans have when collaborating with a robot. Therefore, it is important to analyze and design human-robot collaboration from an interdisciplinary perspective, including robotics, engineering sciences as well as cognitive science, psychology, and ethics~\cite{Inga.2022, londono2022doing}. In \autoref{fig:Figure 15.drawio.png}, we highlight the essential workflow involved in designing and implementing a human-centered approach to robot learning.

We adopt an interdisciplinary perspective to investigate how the perception of robot behavior is influenced by the personal characteristics of the human collaboration partner. The aim of this study is to assess the importance of social robots adjusting their behavior to the individual traits of human collaborators in domestic environments. To achieve this, we designed a case study centered on a classic human-robot collaboration task that involves the joint unpacking of a shopping basket. We assess whether humans perceive the assistance provided by a robot in emptying a shopping basket as efficient and useful, on the one hand, and as comfortable and socially appropriate, on the other. In the present study, the presence of robots in human-inhabited environments is understood as the integration of social agents capable of engaging in high-level social interaction and potentially reshaping our very understanding of what constitutes social interaction. Accordingly, the robot is not merely viewed as a machine, but is instead conceptualized as a social agent within the interactional space. 

In our study, we focused on the design characteristics of robot behavior related to timing, rhythm, and fluency in the interaction. For the robot's design, engineers were involved who were able to assess the level of accuracy of the simulation carried out. We considered four different dimensions of robot behaviors ('antisocial', 'midFluency', 'maxFluency', and 'alternating items') in two different handing conditions ('handover' and 'no-handover'). We varied seven different types of robot behavior (see 3.2) and combined these robot behaviors with characteristics of the human collaboration partner, varying four different human conditions (young female, young male, disabled, and aged). In total, we compared 28 variations of the human-robot collaboration task. We recorded these variations on video and asked participants to assess them from an observer's perspective in an online study. 

Since the type of advanced social robots that perform complex tasks in domestic environments and adapt flexibly to the humans around them are not mature, market-ready products, we had to assume that the participants would have little to no experience with advanced social robots in practice. Furthermore, we also critically questioned how easy or difficult it would be for participants to evaluate the variations of the human-robot collaboration task from an observer's perspective, based solely on video footage, and to relate the robot's behavior to the different human characteristics. To address these concerns, we provided participants with a scenario text at the beginning of the survey that provided information about social robots and emphasized the importance of robot adaptability to different physical conditions of human collaboration partners. 

In addition, we hypothesized that participants would make more differentiated assessments if they had thought more intensively and deeply about the opportunities and challenges of human-robot collaboration and the associated emotions before watching the videos. To foster reflection, we asked half of the participants to create a cognitive-affective map (CAM) on human-robot collaboration before completing the questionnaire survey. Cognitive-affective mapping is a kind of mind mapping technique. Participants create and connect concept nodes on a given topic and additionally assign affective connotations to each concept node (see e.g.,~\cite{Thagard2010, reuter2022direct, fenn2025cognitive}). 

We therefore address two research questions (RQs):
\begin{itemize}
    \item \textbf{RQ-HRC}: How do participants assess a human-robot collaboration (HRC) from the observer's perspective with regard to the combination of robot behavior and human physical preconditions? 
    \item \textbf{RQ-CAM}: Does prior general reflection on human-robot collaboration using cognitive-affective mapping (CAM) lead to differences in the assessment of the specific human-robot collaboration in our study?
\end{itemize}

This study aims to contribute to the field of social robotics from an interdisciplinary perspective, focusing on the development of robots that cater to human preferences and needs within a specific social context. Understanding these factors not only supports the improvement in individuals' willingness to engage with robots, but also shapes their broader perception of robotic technologies~\cite{ahmed2024human}. Furthermore, adopting this socio-technical approach will drive progress in social robotics by establishing a pathway for successful and safe integration, as well as acceptance of robots in everyday social environments.

In the following sections, we provide an overview of related work conducted in the context of human-robot collaboration, human-centered robot design approaches, and cognitive-affective mapping. We provide a comprehensive description of the methodological details of our study and present the results of evaluating our hypotheses. After discussing the results, we end with a conclusion and an outlook on future research.

% \textbfThe Introduction section, of referenced text \cite{bib1} expands on the background of the work (some overlap with the Abstract is acceptable). The introduction should not include subheadings.

% \textbfSpringer Nature does not impose a strict layout as standard however authors are advised to check the indiidual requirements for the journal they are planning to submit to as there may be journal-level preferences. When preparing your text please also be aware that some stylistic choices are not supported in full text XML (publication version), including coloured font. These will not be replicated in the typeset article if it is accepted. 

\section{Related Work}\label{sec2}

In this section, we present relevant studies on human-robot collaboration from both computer science and social sciences. These studies employ quantitative and qualitative methods to analyze various aspects of human-robot collaboration, including efficiency, ethical considerations, and social acceptance. We also present an overview of studies related to human-centered design in AI and robotics, highlighting the importance of these approaches in the development of collaborative robots. 

In particular, we explore how human-centered design contributes to creating more intuitive, accessible, and ethically responsible robotic systems, thereby fostering more natural and efficient interactions between humans and robots in various contexts. Finally, we present an overview of the current state-of-the-art in cognitive-affective mapping, aiming to clarify how this method works and contributes to knowledge production in various fields.

\subsection{Human-Robot Collaboration}\label{subsec2}

Human-robot collaboration (HRC) is an expanding research field with a broad range of applications, emerging future scenarios, and substantial economic potential. It is often regarded as the highest level of interaction, where physical cooperation and shared task execution between humans and robots are enabled \cite{Dzedzickis2024}. HRC-based technologies are expected to play an increasingly significant role across diverse sectors. 

In this context, collaboration refers to the process of joint action, where humans and robots engage in a shared task with the aim of achieving a common goal. Effective human-robot collaboration requires the establishment of a shared plan and mutual understanding between partners. This implies that both the human and the robot must be capable of interpreting the intentions, actions, and goals of the other. For collaboration to be successful, robots must go beyond simple task execution; they need to exhibit cognitive capabilities, including environmental perception, decision-making, planning, learning, and reflective reasoning, features commonly associated with the broader concept of robotic cognition \cite{honerkamp2023n, arce2023padloc, vodisch2023codeps, chisari2024learning}.

Robots generally utilize multi-modal sensor systems, such as vision, lidar, force/torque sensors, and proximity detectors, to continuously perceive and track both their human collaborators and the surrounding environment \cite{li2024safe}. The acquired sensory data is subsequently processed through perception algorithms and sensor fusion techniques to construct a real-time representation of the environmental state and to infer human actions, intentions, and spatial positioning. This perceptual foundation enables adaptive, context-aware behavior, which is essential in dynamic human-robot collaboration scenarios \cite{ajoudani2018progress}.

As robots are increasingly expected to collaborate closely with humans, they are no longer perceived merely as functional tools, but rather as social agents capable of interacting in human-centered environments \cite{londono2024fairness}. Building on this, \cite{janssen2024psychological} argue that human-robot collaboration (HRC) should adopt a psychological need-fulfillment perspective to better support users’ motivation and well-being. They propose the METUX model (Motivation, Engagement, and Thriving in User Experience), which is grounded in Self-Determination Theory (see e.g., \cite{ryan2000self}). This framework emphasizes the fulfillment of three fundamental psychological needs—autonomy, competence, and relatedness—as key to enhancing user engagement and positive experience. By applying this model to various spheres of experience (interface, behavior, and life spheres), the METUX approach offers a holistic framework for the design of socially responsive robots. It reinforces the idea that effective human-robot collaboration must not only be technically robust but also emotionally and psychologically supportive, aligning with users’ intrinsic needs and values. 

\subsection{Human-Centered Approach to Robot Design}\label{subsec21}

A human-centered approach to robot design prioritizes the integration of human needs, behaviors, and values at the core of the design process, particularly in the domain of social robotics. The primary goal is to ensure that robots are developed with a deep understanding of human interaction. Additionally, individual expectations regarding integrating robotic technology into social environments are also carefully considered. Understanding public perceptions and expectations is essential, as it influences both the design and acceptance of these technologies. By aligning robotic functionalities with societal needs and concerns, developers can foster greater trust and smoother integration of robots into everyday social settings, ensuring their effectiveness and ethical deployment.

Several studies have used this approach to examine the impact of social robots on improving interactive and collaborative experiences in different contexts. In~\cite{boschetti2022human}, the authors focus on the human-centered design of industrial robots within a production plant, by presenting a real-time control model for collaborative cells considering human factors, emphasizing the importance of aligning robotic functions with human workflows and needs. The study found that this approach led to increased overall productivity by reducing idle time and minimizing the effort required from operators. Furthermore, the authors observed that a human-centered design improved safety in the workspace, as robots were able to avoid collisions in real time. 

Similarly, \cite{bjorling2019participatory} presents the findings of a study that employed a human-centered design methodology, informed by a participatory approach. This study, conducted over three years, aimed to design and develop social robots to support adolescents' mental health. The participatory process actively engages stakeholders to ensure that robots meet the specific needs and preferences of the target population, especially when designing robots that are expected to collaborate with vulnerable populations. Finally, the authors place these methods in the context of participatory research and recommend a set of principles that may be appropriate for the development of new technologies for vulnerable populations.

Participatory processes are based on the ability of those involved to imagine and think about the possibilities of future technologies. To enable laypeople to engage in participatory processes, scenario texts, for example, can be helpful when the technology is not yet diffused and even prototypes are not yet available \cite{rogers2022maximizing,lee2017steps}. In the current study, for example, we investigate whether cognitive-affective mapping is suitable for promoting laypeople's reflection on the potential opportunities and challenges of social robots.
 
\subsection{Cognitive-Affective Mapping}\label{subsec3}

Cognitive-affective mapping (CAMing) was introduced by the Canadian cognitive scientist and philosopher Paul Thagard~\cite{Thagard2010}. Cognitive-affective mapping aims to capture intricate attitudinal interrelationships to provide essential insights into human decision-making and motivational structures. The method stands in the tradition of other mapping techniques such as cognitive maps or fuzzy cognitive maps~\cite{tolman1948, kosko1986} and was initially intended to visualize existing data sets, e.g., from interviews. Cognitive-affective maps, the result of cognitive-affective mapping, depict attitudes and beliefs in a coherent network. 

Compared to other mapping techniques, the innovation is that cognitive-affective mapping is based on the assumption that emotions influence cognition. Thus, the method allows to add affective ratings to the cognitive content of the network nodes. As a visualization method, cognitive-affective mapping has been used in various areas of research, especially conflict research~\cite{homer2014conceptual}. Recently, cognitive-affective mapping has been further developed into a method that can be used to collect data online~\cite{reuter2022direct, gros2024, fenn2025cognitive}.

As a new method for data collection, cognitive-affective mapping has been explored in different research contexts, including as a method for semantic concept clarification~\cite{nothdurft2021} and for capturing cognitive and affective perceptions of the corona pandemic~\cite{mansell2021}. In \cite{fenn2025cognitive} and \cite{Livanec2022}, cognitive-affective mapping is proposed as a complementary method for technology acceptance research, especially in the context of technologies that are still under development. Cognitive-affective mapping can be used in both affirmative and exploratory designs and provides data that can be analysed quantitatively and qualitatively. \cite{unifreiburg} argues that the method can help bridge the gap between quantitative and qualitative research traditions.
Since cognitive-affective mapping as a data collection method is currently still in the validation phase and its potential has not yet been fully exploited, it is being tested in a wide range of application contexts, including as a tool in doctor-patient communication or as a supportive tool in mediation (for a detailed overview see \cite{unifreiburgCAMResearchInformation}). 

In our study, we investigated whether cognitive-affective mapping can be used as a method of reflection. We conjecture that prior reflection on topics with which participants have little or no exposure in their everyday lives helps participants to evaluate these topics, leading to more pronounced ratings (see 3.5).

\section{Methods}\label{sec3}

In this section, we present a detailed overview of our study by first describing the experimental setup, the manipulated and measured variables, and the sample. This is followed by a presentation of the hypotheses and a description of the statistical models we used.

\subsection{Experimental Setup}\label{subsec24} 

To address RQ-HRC, we conducted a questionnaire study with videos as stimuli. To address RQ-CAM, we combined the questionnaire study with a reflection task using cognitive-affective mapping as a reflection tool. We conducted the study online with two groups: an experimental group that created a cognitive-affective map on human-robot collaboration before completing the questionnaire (CAM group) and a control group that completed the questionnaire only. Thus, we employed a mixed design with the between factor having two levels (CAM group/control group) and the within factor robot behavior having seven levels. Each type of robot behavior was shown in combination with four human characteristics (see Table~\ref{tab:my_label}), resulting in 28 different videos. Each participant rated seven videos, four from the no-handover conditions and three from the handover conditions. Thereby, the four human conditions are assigned counterbalanced according to a Latin square to the four no-handover conditions.

The survey was designed using SoSci Survey\footnote{https://www.soscisurvey.de} and administered via Prolific\footnote{https://www.prolific.com}, where we created two identical experiment descriptions with separate entry points for the CAM group and the control group. This allowed the control group to leave out the CAM part and to participate directly in the questionnaire study. 
The cognitive-affective maps were created with the browser-based software C.A.M.E.L.~\cite{Gouret2022, fenn2025cognitive}. In the following section, we provide a detailed overview of the study design and the methods used for statistical analysis. 

We simulated a human-robot collaboration scenario in which a human and a robot jointly clean a shopping basket in a kitchen environment. The shopping basket contained two categories of products: those that need to be stored in the refrigerator (called "fridge products") and those that do not need to be stored in the refrigerator (called "non-fridge products"). The human's task was to put the fridge products into the refrigerator. The robot was supposed to unpack the shopping basket, pre-sorting the products. It should place the fridge products within the human's reach at the head side of the table to support the human. The non-fridge products should be sorted to the other side. 

\begin{figure}
    \centering
    \includegraphics[width=7.56cm]{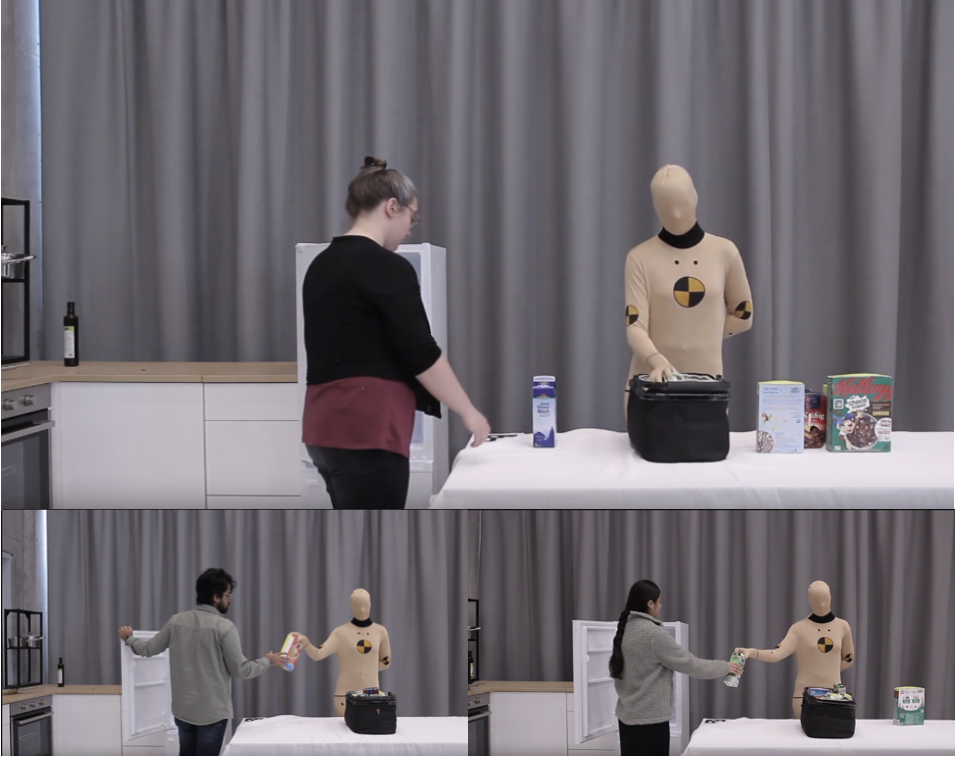}
    \caption{Human-robot collaboration scenario in which a human with different characteristics and a robot jointly clean a shopping basket.}
    \label{fig:hrc.png}
\end{figure}

As the study was intended to provide us with criteria for how to train our robots with social data, we simulated the human-robot collaboration for feasibility reasons by having a human in a morphsuit play the robot (see \autoref{fig:hrc.png}). Our robotics research~\cite{honerkamp2023n} uses a gripper arm that can navigate through space on a mobile platform. Thus, the human imitated these functionalities as closely as possible. The participants of the study were informed that a human was impersonating the robot.%\footnote{We plan to conduct the study with a real robot in the future, allowing for a more accurate examination of this specific human-robot collaboration task}

\subsubsection{Manipulated Variables and Stimulus Materials}\label{subsubsec3}

We created a total of 28 videos showing the human-robot collaboration task as a combination of 7 types of robot behaviors and four different human characteristics. The task that the human and the robot performed together was the same in all videos, but the robot's behavior and the human's characteristics varied.  The four human conditions included female young (FY), male young (MY), disabled male (MD), and aged female (FA). As the videos can be used for imitation learning in robots, we ensured an equal distribution of gender in the conditions in order to avoid gender bias. 

The robot behavior had four dimensions: antisocial (A), maxFluency (F), alternating items (I), and midFluency (M). Three of the four dimensions of robot behavior (maxFluency, midFluency, and alternating items) were examined in two different passing conditions: in a 'handover' and in a 'no-handover' condition.\footnote{In the case of the antisocial robot behavior, we omitted the handover condition because we considered antisocial and handover to be mutually exclusive.} In the 'no-handover' condition, the robot placed the products on the table so that the human collaboration partner had to pick them up. In contrast, in the 'handover' condition, the robot delivered the products directly to the human recipient’s hand. 

The robot behaviors were specified as follows:

\underline{'No-handover' condition:} Antisocial (A): The robot first unpacked all non-fridge products and placed them on its side of the table. Only then did the robot place the fridge products on the table within the reach of the human. Thus, the human had to wait until the robot started to place the fridge products on the table.

MaxFluency (F): The robot first unpacked all the fridge products and placed them on the table within reach of the human. Only then did the robot place the non-fridge products on its side of the table. Thus, the human had many products on their side that might induce stress, especially for the disabled and aged human collaborator.

Alternating items (I): The robot alternately unpacked one fridge product, which it placed on the table within reach of the human, and one non-fridge product, which it placed on its side of the table. Thus, the young female and male had to wait shortly before the next product was placed on the table.

MidFluency (M): The robot alternately unpacked two fridge products, which it placed on the table within reach of the human, and one non-fridge product, which it placed on its side of the table (sorting scheme 2:1). This resulted in optimal unpacking speed for the young female and male while for the disabled and aged collaborator the products were placed somewhat too fast on the table. 

\underline{'Handover' condition:} The robot behaviors maxFluency (F), alternating items (I) and midFluency (M) existed also in a handover condition: maxFluency handover (FH), alternating items handover (IH) and midFluency handover (MH). The respective sorting schemes remained the same, with the difference that the robot did not place the fridge products on the table, but handed them directly to the human.

We assigned the human conditions to the 'no-handover' conditions (4 out of 7 robot behaviors) to the participants using a Latin square design. For the 'handover' conditions, we chose the human condition such that the human characteristic differed from the respective robot behavior in the no-handover condition. Thus, we make sure that each robot's behavior is assigned to each human condition at an equal frequency. This resulted in four combinations, which were counterbalanced over participants (see \autoref{tab:HRCcondition}).\footnote{The videos can be retrieved here: \href{https://rl-fairness.informatik.uni-freiburg.de/study_videos}{study videos}}

\begin{table}
\centering
\caption{Grouping of participants and assessed conditions.}
\setlength{\tabcolsep}{4pt} 
\begin{tabular}{@{}>{\bfseries}c|>{\RaggedRight\arraybackslash}p{0.84\columnwidth}@{}}
\toprule
Group & \textbf{HRC Condition} \\
\midrule
1 & FY-A, MF-F, FA-I, MD-M, FY-FH, MY-IH, FA-MH \\
\midrule
2 & MY-A, FY-F, MD-I, FA-M, MY-FH, FY-IH, MD-MH \\
\midrule
3 & FY-A, MD-F, MY-I, FY-M, FA-FH, MD-IH, MY-MH \\
\midrule
4 & MD-A, FY-F, FY-I, MY-M, MD-FH, FA-IH, FY-MH \\
\bottomrule
\end{tabular}
\label{tab:HRCcondition}
\end{table}

\begin{table*}
    \centering
    \caption{Expected acceptability.}
    \footnotesize
    \begin{tabularx}{\textwidth}{@{}p{2.1cm}|X|X|X|X|X|X|X@{}}
    \toprule%
& \multicolumn{4}{@{}c|@{}}{Robot behavior in the no-handover condition} & \multicolumn{3}{@{}c@{}}{Robot behavior in the handover condition} \\\cmidrule(lr){2-5}\cmidrule(lr){6-8}%
        \textbf{Human condition} & \textbf{Antisocial (A)} & \textbf{Max Fluency (F)} & \textbf{Alternating items (I)} & \textbf{Mid Fluency (M)} & \textbf{Max Fluency (FH)} & \textbf{Alternating items (IH)} & \textbf{Mid Fluency (MH)} \\
    \midrule
        \textbf{Female young (FY)} & Very unacceptable & Medium acceptable & Slightly acceptable & Medium acceptable & Slightly acceptable & Medium acceptable & Acceptable \\
    \midrule
        \textbf{Male young (MY)} & Very unacceptable & Medium acceptable & Slightly acceptable & Medium acceptable & Slightly acceptable & Medium acceptable & Acceptable \\
    \midrule
        \textbf{Disabled (MD)} & Very unacceptable & Unacceptable & Medium acceptable & Slightly acceptable & Unacceptable & Acceptable & Medium acceptable \\
    \midrule
        \textbf{Aged (FA)} & Very unacceptable & Unacceptable & Medium acceptable & Slightly acceptable & Unacceptable & Acceptable & Medium acceptable \\
    \bottomrule
    \end{tabularx}
    \label{tab:my_label}
\end{table*}

Each participant was assigned to one of these four groups and assessed seven out of 28 variations of the human-robot collaboration on a questionnaire. \autoref{tab:my_label} presents an overview of the 28 possible combinations of robot behaviors and human conditions. The table also shows the acceptance ratings of the respective combinations we expected prior to the survey. \footnote{The expected acceptability results range from very unacceptable to unacceptable, medium acceptable, slightly acceptable, and acceptable.} The expected results were formulated by an interdisciplinary team consisting of the authors and two additional researchers representing computer science, robotics, cultural anthropology, and psychology, and were based on the experience with the individual manifestations of the HRC during the production of the videos. 

Before watching the videos and completing the questionnaire, both groups of participants (CAM and control group) were presented with an introductory text that provided general information about current and future application scenarios for socially assistive robots: 

\textit{"Social robots are optimized to interact and collaborate directly with humans. They are designed to support people in everyday tasks. They operate as a teammate and engage in natural interactions with their human partners. In other words, social robots mimic human behavior to a certain extent. Here are some exemplary areas of application for social robots: In public buildings, such as airports or museums, social robots can navigate dynamic crowds and guide people on their way. They can also provide receptionist services. In the healthcare sector, they support caregivers when handing out medication or disinfecting surfaces. At the workplace, they can collaborate directly with humans, e.g., by helping to carry heavy loads or assemble a product.  Social robots are also used more and more in domestic environments for various purposes. They can help with housework, e.g., they can support ironing, tidying, or cooking. Due to their sociability, they are also deployed, among others, to provide companionship to elderly, lonely people, to support children in learning cognitive and emotional skills, or to train conflictual conversations with managers. As they are designed to operate in close proximity to humans and since they are equipped with a variety of sensors, cameras, microphones, and artificial intelligence, they have to meet high requirements both in terms of their motion patterns and their social skills, as well as their privacy policy. Broadly speaking, they must ensure both the safety of the individuals with whom they interact and human comfort."}

After the introductory text, the experimental group (CAM group) created a CAM on the topic of human-robot collaboration, paying particular attention to the fact that humans have different needs (e.g., due to disability or aging). The instruction for the CAM group was: \textit{"What comes to your mind when you think of social robots that work directly with humans? Which conditions must be met for the human-robot collaboration to be successful, comfortable, and safe? What has to be avoided? How would you describe the relationship between humans and social robots? Which feelings do you have about it? Please consider also that humans might have specific needs, e.g., handicapped or elderly people. We now ask you to create a Cognitive-Affective Map (CAM), a type of mind map, for human-robot collaboration. With a CAM, you can visualize your attitudes and feelings about the topic."} 

\begin{figure}
    \centering
    \includegraphics[width=7.56cm]{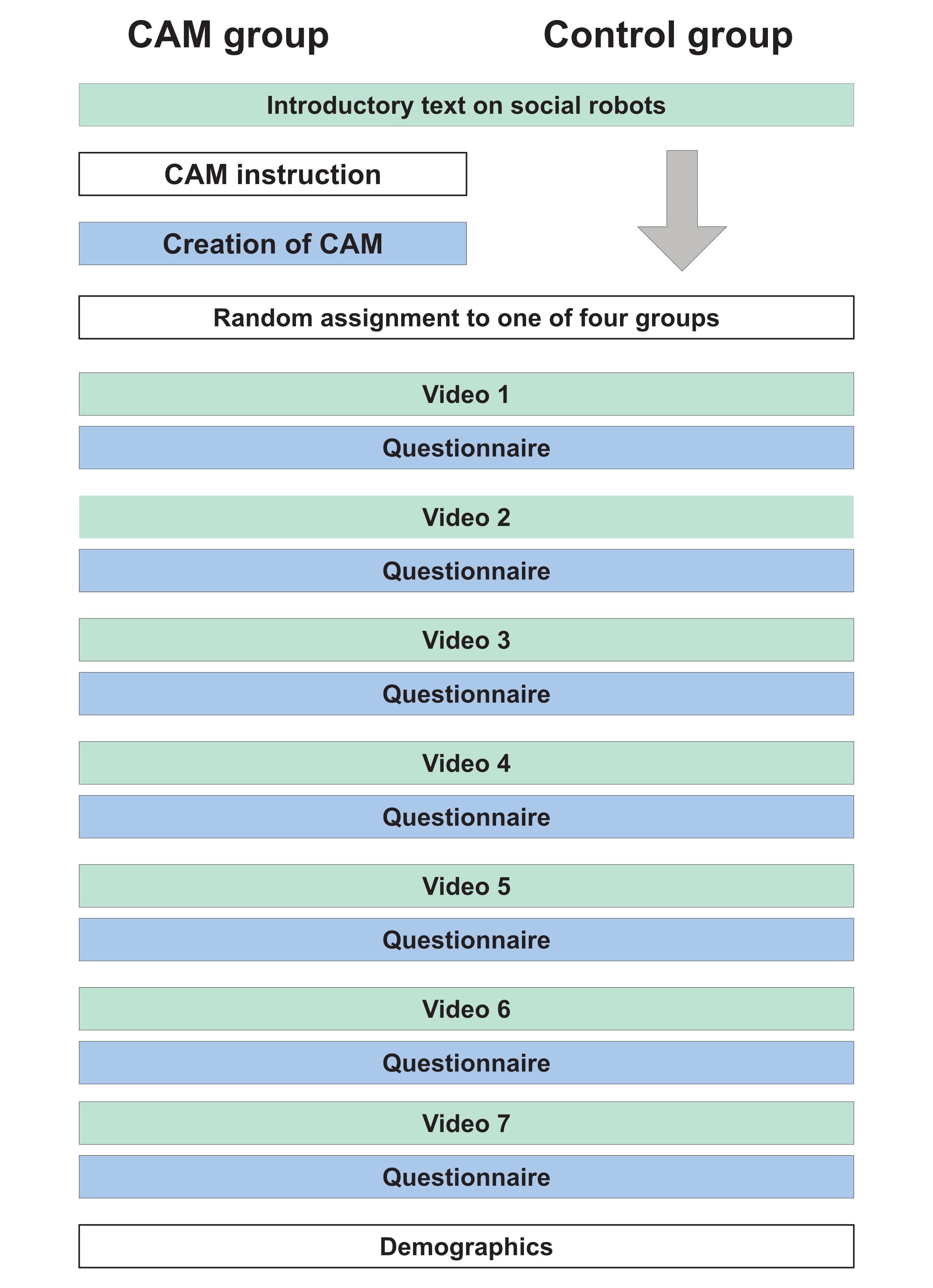}
    \caption{Schematic overview of the experiment procedure. Random assignment to one of four groups according to \autoref{tab:HRCcondition}.}
    \label{fig:Graphics_procedure}
\end{figure}

Following this, the participants were explained how to create a cognitive-affective map using the browser-based software C.A.M.E.L.~\cite{Gouret2022}.
For a schematic overview of the experiment procedure, see \autoref{fig:Graphics_procedure}.

\subsection{Measurements}\label{subsec4}

Our two outcome variables were: a) the assessment of the human-robot collaboration, and b) the difference in ratings between the participants in the CAM group and the control group.

\subsubsection{Questionnaire}

\begin{figure}
    \centering
    \includegraphics[width=7.56cm]{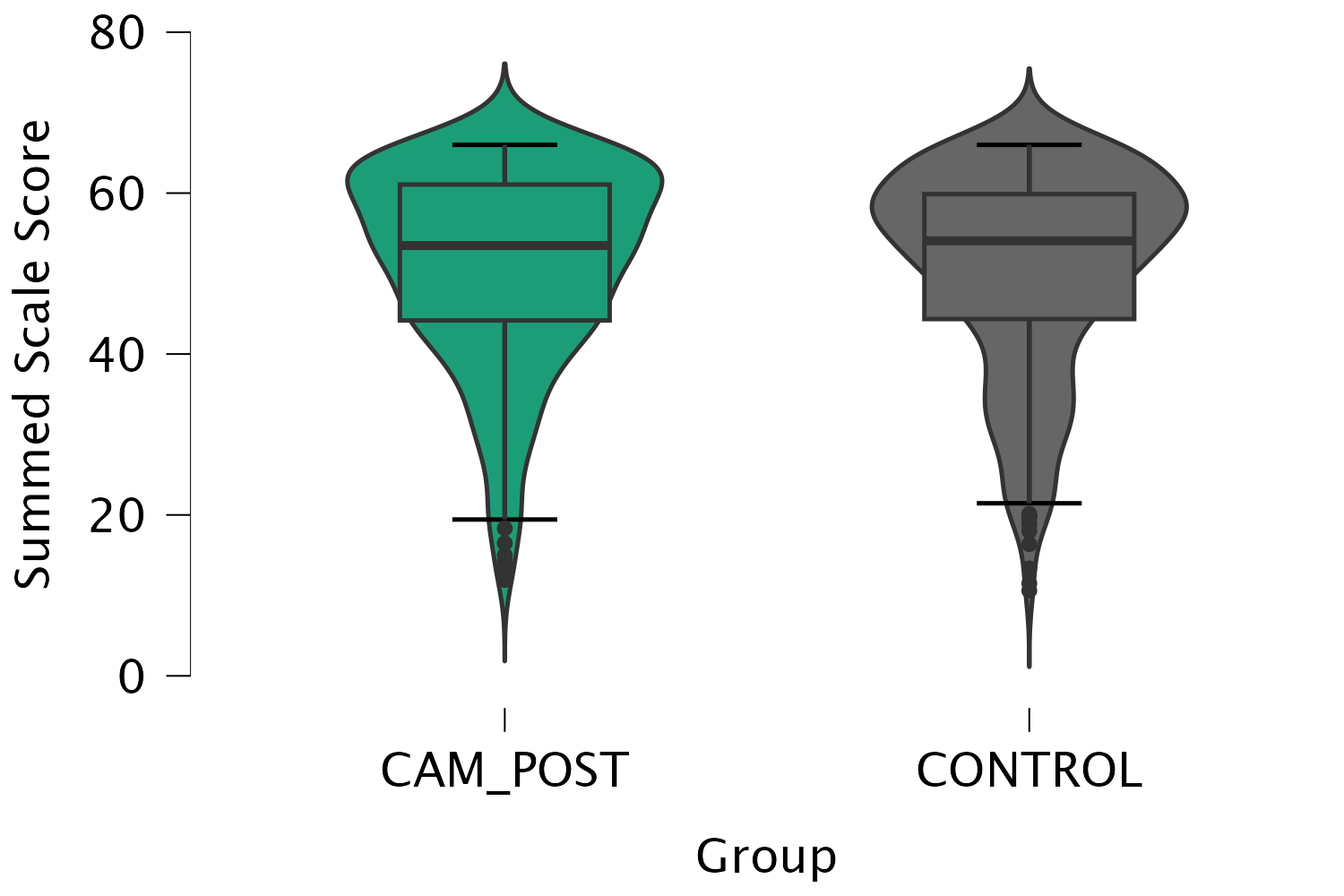}
    \caption{Violin and box plots of the summed scale scores. The distribution of the overall assessment of human-robot collaboration for the CAM group (green) and the control group (gray) is shown. The boxplots show the median, the first and third quartile; the violin plots show the density distribution of the data points.}
    \label{fig:Boxplot_summed_scale_score}
\end{figure}

For RQ-HRC, the assessment was measured using 7-point Likert subscales for Procedural Fairness ~\cite{Ting2013, Carr2007, Sindhav2006, Leventhal1980}, Interpersonal Fairness~\cite{Ting2013, Carr2007, Bies1986, Sindhav2006}, Perceived Usefulness~\cite{Davis1989}, Quality of Interaction~\cite{baraglia2016initiative}, System Performance~\cite{baraglia2016initiative}, Positive Teammate Traits~\cite{Hoffman2019}, Attitude Towards Using the Robot~\cite{Venkatesh2003}, as well as GODSPEED III: Likability and GODSPEED IV: Perceived Intelligence~\cite{Bartneck2009}.   Additionally, a General Rating was included (see \autoref{tab:scales_tab}). We computed a summed scale score to obtain an overall assessment of the human-robot collaboration. A high value in the summed scale score corresponds to an overall positive assessment of the human-robot collaboration; consequently, high values in \autoref{fig:Boxplot_summed_scale_score} reflect a positive perception of collaboration.

It is important to note that several items from the referenced subscales were adapted in the questionnaire by replacing general terms such as “technological device” or “technology” with the more specific term “robot”. This modification was made to avoid ambiguity and to ensure that participants clearly understood that the questions referred explicitly to robotic systems. Furthermore, since this study conceptualizes robots in social environments not merely as mechanical tools, but as social agents, and given that our aim was to explore how social robots can adapt to the characteristics of human collaboration partners in domestic settings, we selected subscales that capture both functional-cognitive dimensions (Perceived Usefulness, System Performance, Quality of Interaction) and affective-human-values-related dimensions (Procedural Fairness, Interpersonal Fairness, Positive Teammate Traits, Attitude Toward Using the Robot, GODSPEED III: Likeability, and GODSPEED IV: Perceived Intelligence). 

These subscales are able to capture nuanced human perceptions of a robot’s sociability, relational behavior, and social presence, which are central to evaluating meaningful human-robot collaboration. We chose these subscales in particular for the following reasons:
As socially assistive robots must design their processes in such a way that they are equally accessible and appropriate for all user groups without disadvantaging certain groups, procedural fairness makes a decisive contribution to HRC being perceived as successful, especially if humans with protected characteristics are involved. Interpersonal fairness focuses on the quality of interpersonal treatment during formal decision-making processes. Respect, appropriateness, and the avoidance of bias are particularly relevant when robots interact with people with different abilities and needs. 
The assessment of the perceived usefulness of the robot is fundamental to acceptance and intention to use. According to \cite{Davis1989}, perceived usefulness is a key factor in technology acceptance. For people with protected characteristics, the definition of usefulness may vary—for aged people or people with disabilities, this may mean that the robot takes over or supports specific tasks that are difficult for them.

The quality of interaction subscale measures the smooth and effective interaction between humans and robots. High-quality interaction is significant for users with special needs. \cite{Hoffman2019} emphasizes the importance of fluency in human-robot collaboration, which can be especially crucial for people with disabilities or the elderly, as they may require more time or specific adaptations.
System performance refers to the technical effectiveness and efficiency of the robot. A system designed to adapt to protected characteristics must deliver consistent performance that meets the user's specific requirements. Evaluating system performance allows us to understand whether the robot meets the technical requirements necessary to support different user groups.

The Positive Teammate Traits subscale measures the extent to which the robot is perceived as a positive and supportive teammate. When working with humans with different abilities and protected characteristics, it is crucial that the robot is seen as a helpful partner that adapts to individual needs. This dimension is particularly relevant as it reflects the social acceptance and integration of the robot into everyday life.
Attitude Towards Using the Robot is based on the Technology Acceptance Model~\cite{Davis1989} and measures the general attitude of users towards the robot. Attitude can be strongly influenced by how well the robot adapts to specific needs, especially for people with protected characteristics. \cite{Venkatesh2003}~extended this model to a unified view of technology acceptance, which is also relevant for our study.

The GODSPEED III: Likeability and GODSPEED IV: Perceived Intelligence subscales from Bartneck et al. \cite{Bartneck2009}~measure how likeable and intelligent the robot is perceived to be. Our literature review of studies on HRC shows that likeability and intelligence are decisive factors for user acceptance. For people with protected characteristics, the robot must be not only perceived as intelligent enough to recognize individual needs, but also as socially comfortable to promote long-term acceptance.

\subsubsection{Cognitive-Affective Mapping (CAM)}

\begin{table*}
\centering
\caption{Overview of the scales used in the user study. For each item, participants indicated their agreement on a 7-point Likert scale. For the sum score, negative items were recoded.}
\footnotesize
\begin{tabularx}{\textwidth}{p{3.1cm}p{3.6cm}p{1.8cm}X}
\toprule
\textbf{Dimension} & \textbf{Criteria} & \textbf{Type} & \textbf{Items} \\
\midrule
Procedural Fairness (PF)~\cite{Ting2013, Carr2007, Sindhav2006, Leventhal1980} & Perceived fairness in the robot's decision-making process & Affective & 1. The process of working with the robot was generally fair. 2. The activities of the robot were conducted without bias. 3. The robot attempted to meet the human’s needs. 4. The time that the human had to wait for the robot's actions and response was appropriate. \\
\midrule
Interpersonal Fairness (IF)~\cite{Ting2013, Carr2007, Bies1986, Sindhav2006} & Respectfulness and fairness in robot-mediated interactions & Affective & 1. The robot seemed to be polite. 2. The robot seemed to be respectful. 3. The robot treated the human with dignity. 4. The robot seemed to be courteous. 5. The robot seemed to be friendly. 6. The robot treated the human with an unbiased attitude. \\
\midrule
Perceived Usefulness (PU)~\cite{Davis1989} & Belief that using the robot improves task performance & Cognitive & 1. Using the robot in my home would enable me to accomplish tasks more quickly. 2. Using the robot would improve my performance. 3. Using the robot at home would increase my productivity. 4. Using the robot would enhance my effectiveness at home. 5. Using the robot would make it easier to do my housework. 6. I would find the robot useful at home. \\
\midrule
Quality of Interaction (QI)~\cite{baraglia2016initiative} & Perceived quality of collaboration with the robot & Cognitive & 1. The robot was helpful in accomplishing the task. 2. The collaboration appeared natural. 3. The robot and the human worked efficiently together. 4. The robot and the human worked fluently together. 5. The robot and the human contributed equally to the completion of the task. \\
\midrule
System Performance (SP)~\cite{baraglia2016initiative} & Perception of technical functionality of the robot & Cognitive & 1. The robot was able to accurately perceive the human’s actions. 2. The robot was able to keep track of the task progress. \\
\midrule
Positive Teammate Traits (PT)~\cite{Hoffman2019} & Perception of the robot as a good teammate & Affective & 1. The robot seemed to be intelligent. 2. The robot seemed to be trustworthy. 3. The robot seemed to be committed to the task. \\
\midrule
Attitudes Toward Using the Robot~\cite{Venkatesh2003} & Overall response to the idea of using the robot & Affective & 1. Using the robot is a good idea. 2. Using the robot is a bad idea. 3. The robot makes housework more interesting. 4. Collaborating with the robot would be fun. 5. I would like to collaborate with the robot. \\
\midrule
Godspeed III: Likeability~\cite{Bartneck2009} & How likable or socially appealing the robot is & Affective & 1. Dislike–Like. 2. Unfriendly–Friendly. 3. Unkind–Kind. 4. Unpleasant–Pleasant. \\
\midrule
Godspeed IV: Perceived Intelligence~\cite{Bartneck2009} & How intelligent and competent the robot seems & Affective & 1. Incompetent–Competent. 2. Ignorant–Knowledgeable. 3. Irresponsible–Responsible. 4. Unintelligent–Intelligent. 5. Foolish–Sensible. \\
\midrule
General Rating & Overall evaluation of human-robot collaboration & -- & 1. Overall, I liked the way the human and the robot collaborated. 2. Overall, the robot has adapted adequately to the needs of the human. \\
\bottomrule
\end{tabularx}
\label{tab:scales_tab}
\end{table*}

The cognitive-affective maps that we collected for RQ-CAM function as independent variables in this study to investigate the potential of cognitive-affective mapping as a reflection tool. We measured whether the ratings of the CAM group on the summed score of the scales listed in \autoref{tab:scales_tab} differed significantly from those of the control group. 
In a subsequent study, we will examine the cognitive-affective maps as a dependent variable and analyze them according to quantitative and qualitative standards. 

\subsection{Power Analysis and Participants}\label{subsec41}
112 participants (56 each in the CAM and control groups and 14 per group according to the Latin square design) took part in the experiment. We determined the sample size for the between subject comparison CAM/control based on G*Power. For a medium effect size (d=0.5) in a one-tailed t-test with a standard .05 alpha error probability, we needed 51 participants per group to obtain a power of 0.8. Due to counterbalancing reasons, we collected data from 56 participants per group. This sample size per group (N=56) resulted in a power of 0.99 for the $2\times2$ ANOVA with the between subject factor group and the respective within subject comparisons with standard .05 alpha error probability when assuming medium effect sizes for the HRC-related hypotheses (correlation among repeated measures = .5 nonsphericity correction).

Participants were recruited on prolific.co whereby we selected participants who are located in the US and are fluent in English. Participation was compensated with £ 11 per hour, which is approximately equivalent to the German minimum wage. 
To reach the target sample size of 56 participants per group, we finally had to recruit 61 participants in the control group (dropout rate: 8\%) and 72 participants in the CAM group (dropout rate: 22\%).

\subsection{CAM Hypotheses}\label{subsec5}

To address \textbf{RQ-CAM}, we tested the following CAM-related hypotheses: 

\textbf{H1}: Reflecting on human-robot collaboration by drawing a CAM before completing a questionnaire leads to more pronounced ratings regarding the subscales that relate to human values and affective aspects (Procedural fairness, Interpersonal fairness, Positive teammate traits, Attitude towards using the robot, GODSPEED III and IV) rather than those that refer more to functionality and cognitive aspects (Perceived usefulness, System performance, Quality of interaction). 

That is \textbf{H1.1} Antisocial is rated worse for CAM drawers compared to non-CAM drawers, and \textbf{H1.2} the acceptable conditions (maxFluency, midFluency, alternating items) are rated more acceptable by CAM drawers compared to non-CAM drawers.
Additionally, we analyzed whether drawing a CAM has an impact on the assessment of each type of robot behavior in combination with the human characteristics. Please note that this analysis is exploratory and intended to formulate hypotheses for future research.  

\subsection{HRC Hypotheses}\label{subsec6}

Regarding \textbf{RQ-HRC}, we predicted acceptability ratings as shown in \autoref{tab:my_label} and derived the following hypotheses: 

\textbf{H2}: The antisocial condition is rated worse than all other conditions.

\textbf{H3}: In the 'no-handover' conditions, midFluency (M) and maxFluency (F) are rated better than alternating items and antisocial for young humans (FY and MY).

\textbf{H4}: In the 'no-handover' conditions, maxFluency is rated worse than alternating items for disabled and aged humans (D and A).

\textbf{H5}: In the 'no-handover' condition, midFluency (M) is rated worse than alternating items (I) for disabled and aged humans (D and A).

\textbf{H6}: Across all human conditions (FY, MY, D, and A), alternating items (I) and midFluency (M) are rated better in the 'handover' conditions than in the 'no-handover' conditions.

\textbf{H7}: Across all human conditions (FY, MY, D, and A), maxFluency (F) is rated worse than the alternating items (I) and midFluency (M) conditions in the 'no-handover' condition.

\textbf{H8}: In the young human conditions (FY and MY), maxFluency (F) is rated worse in the 'handover' condition than in the 'no-handover' condition.

\textbf{H9}: In disabled and aged humans (D and A), maxFluency (F) in the handover conditions is equally negatively rated as maxFluency (FH) in the 'no-handover' conditions.

In addition, we exploratorily analyzed whether handover and no-handover were rated significantly differently across all conditions. 

\subsection{Statistical Models}\label{subsec23}

As a data basis for the analyses with regard to RQ-CAM and RQ-HRC, we calculated the sum of the ratings on the individual scales and summed the scale values to an overall value. 

For the analysis of the CAM-related hypotheses (H1, H1.1, H1.2), we conducted independent samples t-tests. We tested whether the ratings of the CAM group differed from the control group on the scales that are more related to affective aspects, as well as overall across the sum of all scales. For the HRC-related hypotheses (H2–H8), we performed 2x2 ANOVAs, analyzing between subject factors by group and within subject comparisons according to the respective hypothesis. Additionally, we examined potential interactions between the group and the relevant conditions. For H9, we performed a $2\times2$ Bayes ANOVA, as we expected equal ratings between conditions. For the exploratory analysis of the handover/no-handover comparison, we used a paired samples t-test.

\section{Results}\label{sec6}

In the following, we present the statistical analysis (t-test) results for hypotheses H1, H1.1, and H1.2, which proposed that drawing a CAM influences the affective and cognitive subscales. These hypotheses examine how the creation of a CAM impacts participants' responses on these specific dimensions, focusing on emotional and cognitive responses. Following this, we present the results of the statistical analyses (ANOVA and Bayesian ANOVA) for the hypotheses related to human-robot collaboration (H2–H9). These analyses investigate the effects of different robot behaviors and human conditions on the participants’ perception of collaboration, highlighting any significant differences and interactions between groups and conditions.

\subsection{CAM-related Results}\label{subsec22}

Reflecting on human-robot collaboration by drawing a CAM before completing the questionnaire had no significant influence on the ratings on the affective or cognitive subscales (\autoref{tab:scales_tab}). This contradicts our original assumption that the CAM group would show more pronounced ratings in the affective subscales. 
Moreover, for \textbf{H1.1}, the data show no significant difference in the evaluation of 'antisocial' robot behavior between the CAM and the control groups. Similarly, we found no evidence in the data that the CAM group would rate the non-antisocial, i.e., the basically acceptable robot behaviors ('midFluency', 'maxFluency' and 'alternating') more positively than the control group, so that we also have to reject \textbf{H1.2}.

\subsection{HRC-related Results}\label{subsec7}

\begin{table*}
    \centering
    \footnotesize
    \caption{Overview of hypotheses testing depending on the personCharacteristics.}
    \label{tab:results_overview}
    \begin{tabularx}{\textwidth}{llX X l}
        \toprule
        \textbf{Hypothesis} & \textbf{PersonCharacteristics} & \textbf{Compared robot behaviors} & \textbf{Results} & \textbf{Significance} \\
        \midrule
        H2 & all & antisocial vs. all others & antisocial was rated significantly worse than all other behaviors. No significant differences were found between the non-antisocial conditions. & confirmed \\
        H3 & young (FY, MY) & midFluency, maxFluency vs. antisocial, alternating & midFluency and maxFluency significantly higher rated than antisocial; only midFluency significantly higher than alternating (p = .042) & partially confirmed \\
        H3 & aged (A) & midFluency, maxFluency vs. antisocial, alternating & Only midFluency significantly higher rated than antisocial; no differences to alternating & partially confirmed \\
        H4 & disabled, aged (D, A) & maxFluency vs. alternating (no-handover) & maxFluency not lower rated than alternating; no significant difference & not confirmed \\
        H5 & disabled, aged (D, A) & midFluency vs. alternating (no-handover) & midFluency not lower rated than alternating; no significant difference & not confirmed \\
        H6 & all & alternating, midFluency: handover vs. no-handover & Only alternating was rated significantly higher in handover than in no-handover ($\omega^2$ = 0.036); midFluency no significant difference & partially confirmed \\
        H7 & all & maxFluency vs. alternating, midFluency (no-handover) & maxFluency was not rated worse than alternating or midFluency in no-handover & not confirmed \\
        H8 & young (FY, MY) & maxFluency: handover vs. no-handover & maxFluency was rated not worse in the handover than in the no-handover & not confirmed \\
        H9 & disabled, aged (D, A) & maxFluency: handover vs. no-handover & Rating of maxFluency in handover similar to no-handover (Bayes Factor $BF_{10}$ = 0.838, P(M|data) = 54.4\%) & partially confirmed \\
        H9 & aged (A) & maxFluency: handover vs. no-handover & Rating of maxFluency in handover similar to no-handover (Bayes Factor $BF_{10}$ = 0.516, P(M|data) = 66\%; stronger evidence for the null model) & confirmed \\
        \bottomrule
    \end{tabularx}
\end{table*}

Given the non-significant differences between the CAM group and the control group in the assessment of the human-robot collaboration (HRC), we decided to aggregate the data from both groups for further analysis. With this approach, we aim at a more comprehensive and statistically robust analysis of the perception and evaluation of the different conditions of the HRC presented in our experiment, regardless of the initial group assignment. An overview of the results for the HRC-related hypotheses is given in \autoref{tab:results_overview}.

\begin{figure}
    \centering
    \includegraphics[width=7.56cm]{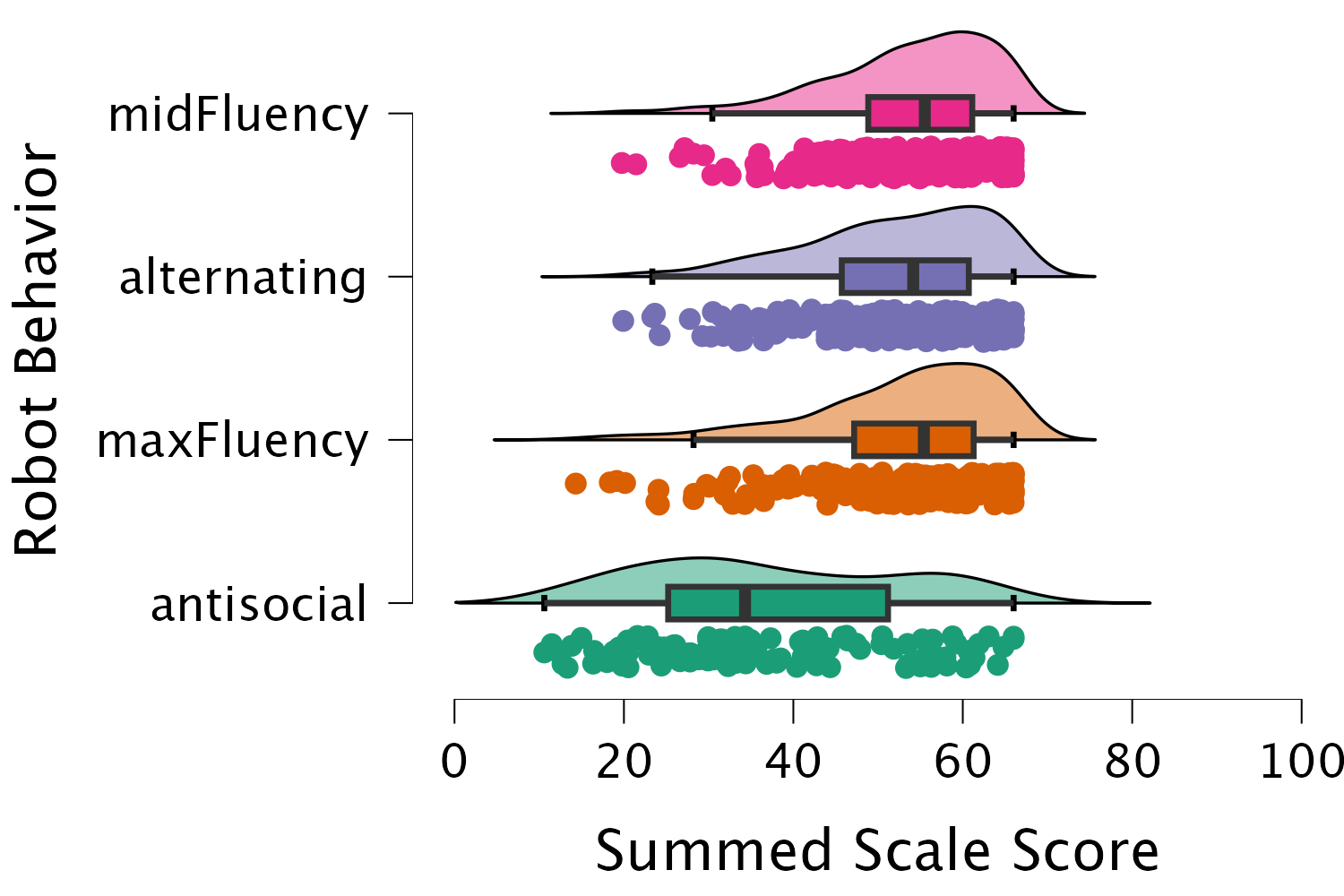}
    \caption{H2: Distribution across the different robot behaviors. The antisocial condition is rated worse than all other conditions.}
    \label{fig:Raincloud_H2}
%    \vspace{-0.4cm}
\end{figure}

Regarding \textbf{H2}, a Welch ANOVA revealed a statistically significant difference in the ratings of the different robot behaviors with F(3, 349.070) = 37.975, p $<$ .001, and $\omega$² = 0.196. The effect size ($\omega$² = 0.196) indicates a significant effect, which underlines the practical significance of the differences found. To specifically confirm that antisocial behavior is rated lower than all other behaviors, we conducted Games-Howell post-hoc tests. These compare pairwise between antisocial behavior and each of the other behavior types, allowing a more detailed analysis of the differences.
\autoref{tab:games-HowellPostHocComparisons-FluencyH2} shows that H2 was fully confirmed. The antisocial robot behavior was rated consistently and significantly worse than all other behaviors, while no significant differences were found between the non-antisocial behaviors (\autoref{fig:Raincloud_H2}).

\begin{table*}
\centering
\footnotesize
\caption{H2: Post-hoc comparisons of the fluency conditions. Antisocial is rated worse than all other conditions.}
\label{tab:games-HowellPostHocComparisons-FluencyH2}
\begin{tabular}{lrrrrrrrr}
\toprule
 & \textbf{Mean} & \multicolumn{2}{c}{\textbf{95\% CI for}}& & & &  \\
\textbf{Comparison} & \textbf{Difference} & \multicolumn{2}{c}{\textbf{Mean Difference}} & \textbf{SE} & \textbf{t} & \textbf{df} & \textbf{p$_\text{tukey}$} \\
\midrule
 &  & \textbf{Lower} & \textbf{Upper} &  &  & &  \\
antisocial - maxFluency & $-15.808$ & $-20.036$ & $-11.580$ & $1.629$ & $-9.701$ & $168.158$ & $<$ .001 \\
antisocial - alternating & $-15.141$ & $-19.332$ & $-10.950$ & $1.615$ & $-9.378$ & $162.994$ & $<$ .001 \\
antisocial - midFluency & $-16.687$ & $-20.815$ & $-12.558$ & $1.590$ & $-10.496$ & $154.404$ & $<$ .001 \\
maxFluency - alternating & $0.667$ & $-1.957$ & $3.291$ & $1.018$ & $0.655$ & $441.943$ & $0.914$ \\
maxFluency - midFluency & $-0.879$ & $-3.400$ & $1.643$ & $0.978$ & $-0.899$ & $434.959$ & $0.806$ \\
alternating - midFluency & $-1.546$ & $-4.003$ & $0.912$ & $0.953$ & $-1.622$ & $440.618$ & $0.367$ \\
\bottomrule
\multicolumn{8}{l}{\textit{Note.} CI = Confidence Interval; df = degrees of freedom.}
\end{tabular}
\end{table*}

\begin{figure}
    \centering
    \includegraphics[width=7.56cm]{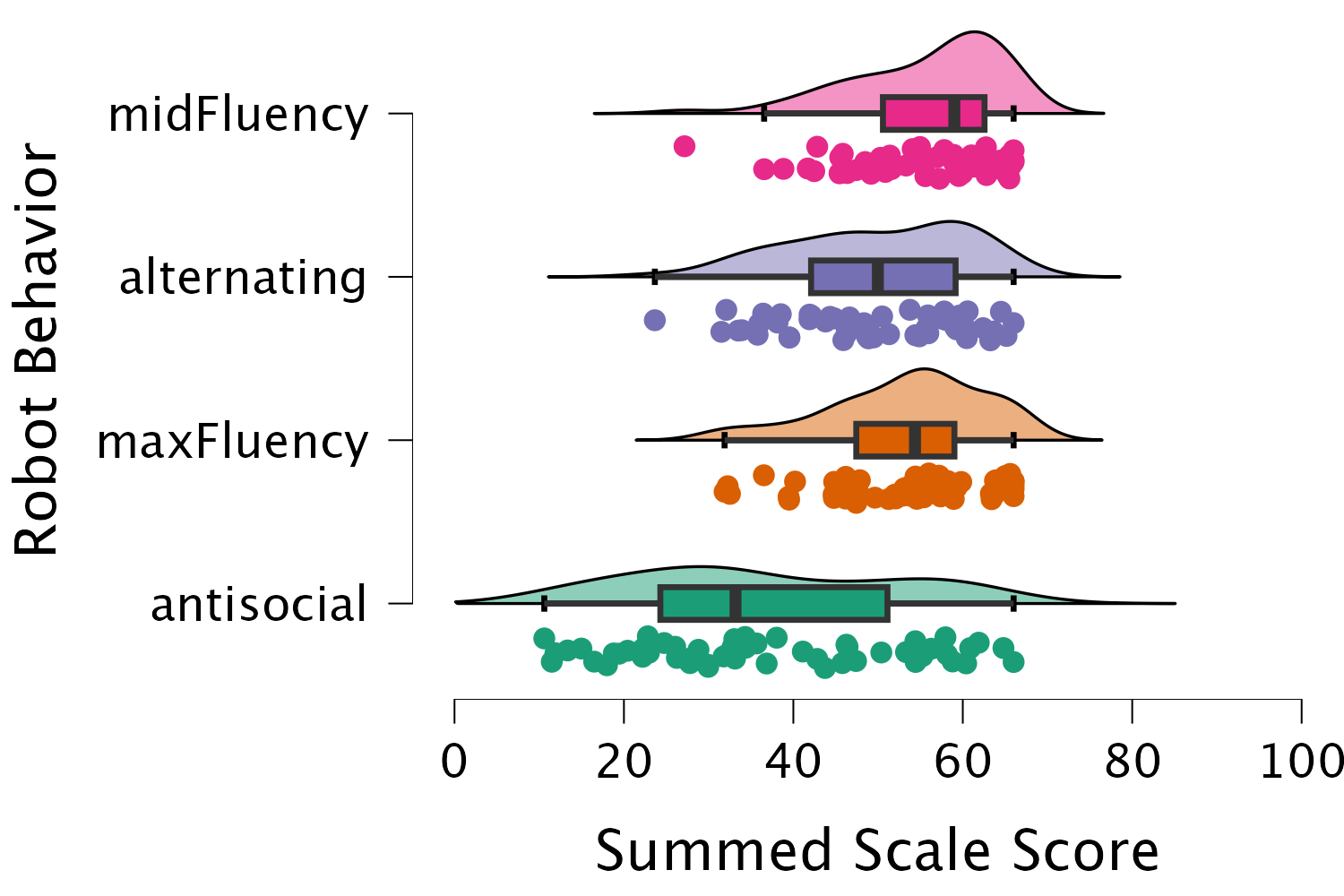}
    \caption{H3 for personCharacteristics = young: Comparison of 'midFluency' and 'maxFluency' on the one hand and 'antisocial' and 'alternating' on the other in the no-handover conditions.}
    \label{fig:Raincloud_H3_young}
%    \vspace{-0.4cm}
\end{figure}

\textbf{H3} which assumed that in the no-handover conditions the robot behaviors 'midFluency' and 'maxFluency' are rated more positively than 'antisocial' and 'alternating' when the collaborating human has no protected characteristics (personCharacteristics = young; H3) was partially confirmed (see \autoref{fig:Raincloud_H3_young}). 'MidFluency' and 'maxFluency' were both rated significantly more positively than 'antisocial'. Compared to 'alternating', however, the Games-Howell post-hoc test with Tukey correction showed that only 'midFluency' was rated slightly more positively with a p-value of .042 (\autoref{tab:games-HowellPostHocComparisons-FluencyH2}).

\textbf{H4} and \textbf{H5} have to be rejected. If the human has protected characteristics ($personCharacteristics\neq young$), participants rate neither 'midFluency' nor 'maxFluency' more negatively than the 'alternating' behavior in the no-handover conditions. 

For hypothesis \textbf{H6} that 'midFluency' and 'alternating' are rated higher in the handover conditions than in the no-handover conditions across all personCharacteristics, we find mixed evidence in the data. 
The one-way ANOVA revealed a significant main effect of the handover condition on the ratings of 'midFluency' and 'alternating' with F(1, 444) = 8.632, p = .003 and $\omega$² = 0.017, 95\% CI [0.001, 0.048]. The effect size ($\omega$² = 0.017) indicates a small effect that may still have meaningful implications in real-world settings. If we consider the robot behaviors 'midFluency' and 'alternating' separately, we find that only 'alternating' is rated significantly more positively in the handover conditions than in the no-handover conditions, with an effect size of $\omega$² = 0.036. We could not find any supporting hints in the data for our two 'maxFluency'-related hypotheses \textbf{H7} and \textbf{H8}. 

The Bayesian ANOVA, which we used to test \textbf{H9}, reveals inconclusive evidence regarding differences in the assessment of maxFluency between the handover and no-handover conditions for disabled and aged humans ($BF_{10}$ = 0.838, P(M|data) for null model = 54.4\%). Therefore, the hypothesis that both conditions are evaluated equally can not be refuted, but the evidence is also not strong enough to confirm it unequivocally.
Analyzing personCharacteristics=aged separately, the Bayes-ANOVA shows moderate evidence that the evaluation of maxFluency is similar in the handover and no-handover conditions ($BF_{10}$ = 0.516, P(M|data) for null model = 66.0\%). Therefore, the hypothesis that both conditions are evaluated equally can not be refuted, and the data even slightly support this assumption.

\subsubsection{HRC-related Exploratory Analysis}

\begin{figure}
    \centering
    \includegraphics[width=7.56cm]{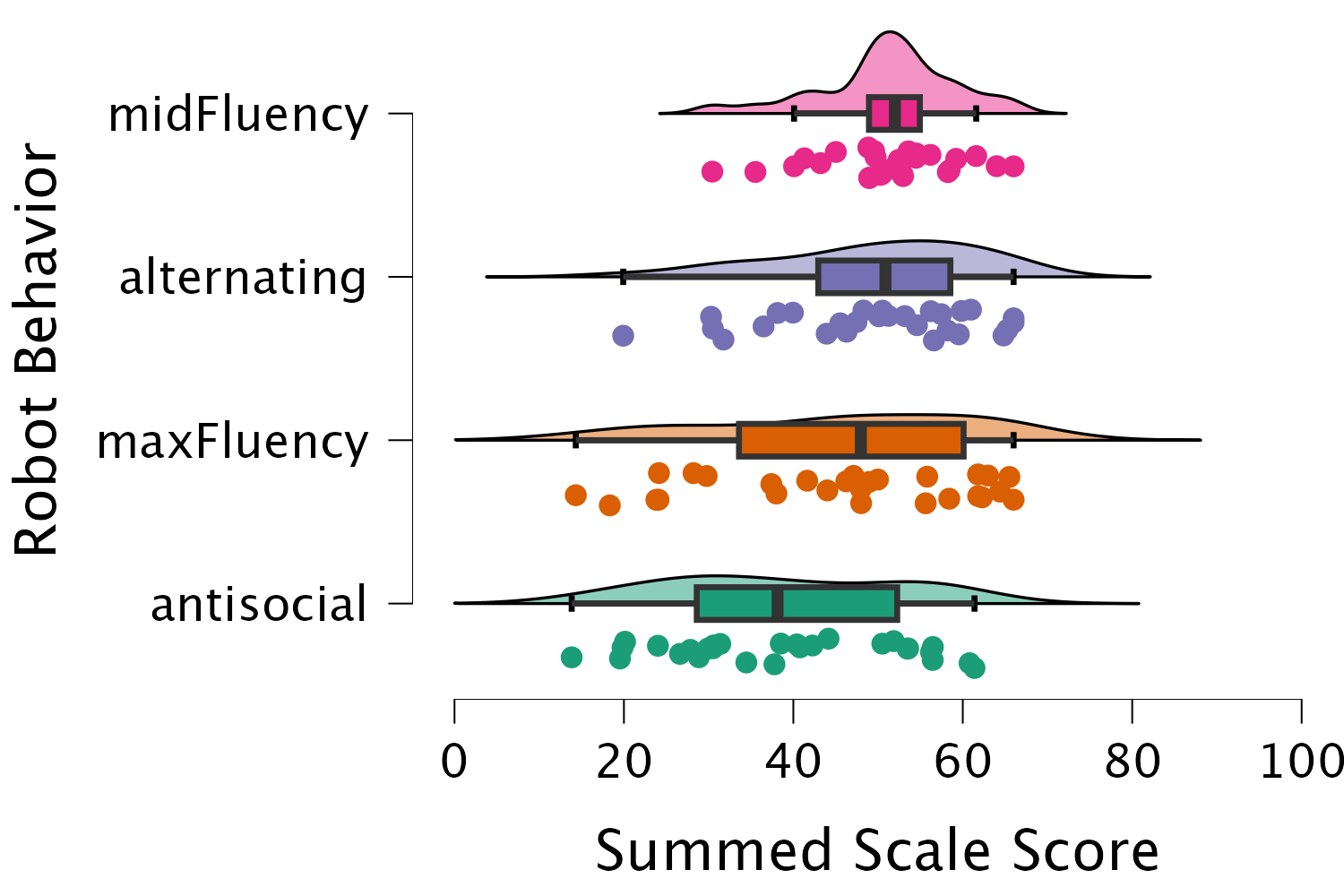}
    \caption{H3 for personCharacteristics = aged.}
    \label{fig:Raincloud_H3_aged}
%    \vspace{-0.4cm}
\end{figure}

We exploratively analyzed H3 under the condition that the collaborating person has the protected characteristic ‘aged’ (personCharacteristics = aged). For this condition, a completely different picture is revealed. Here, only the 'midFluency' behavior is rated significantly more positive than 'antisocial', while there is no evidence for differences between 'midFluency' and 'maxFluency' compared to 'alternating' (see \autoref{fig:Raincloud_H3_aged}).

\begin{figure}
    \centering
    \includegraphics[width=7.56cm]{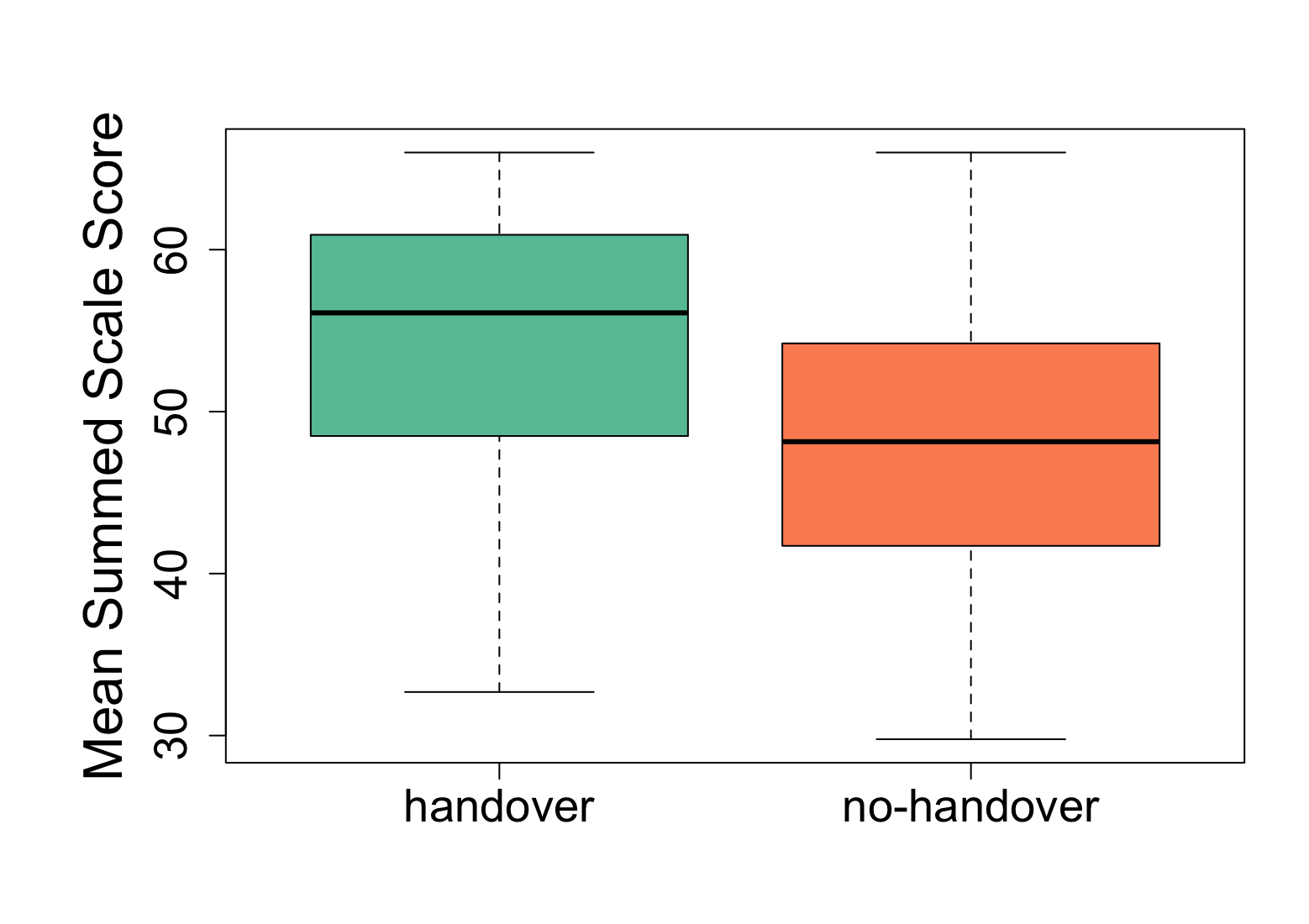}
    \caption{Exploratory comparison of handover and no-handover for personCharacteristics = aged.}
    \label{handover_boxplot}
%    \vspace{-0.4cm}
\end{figure}

\begin{figure}
    \centering
    \includegraphics[width=7.56cm]{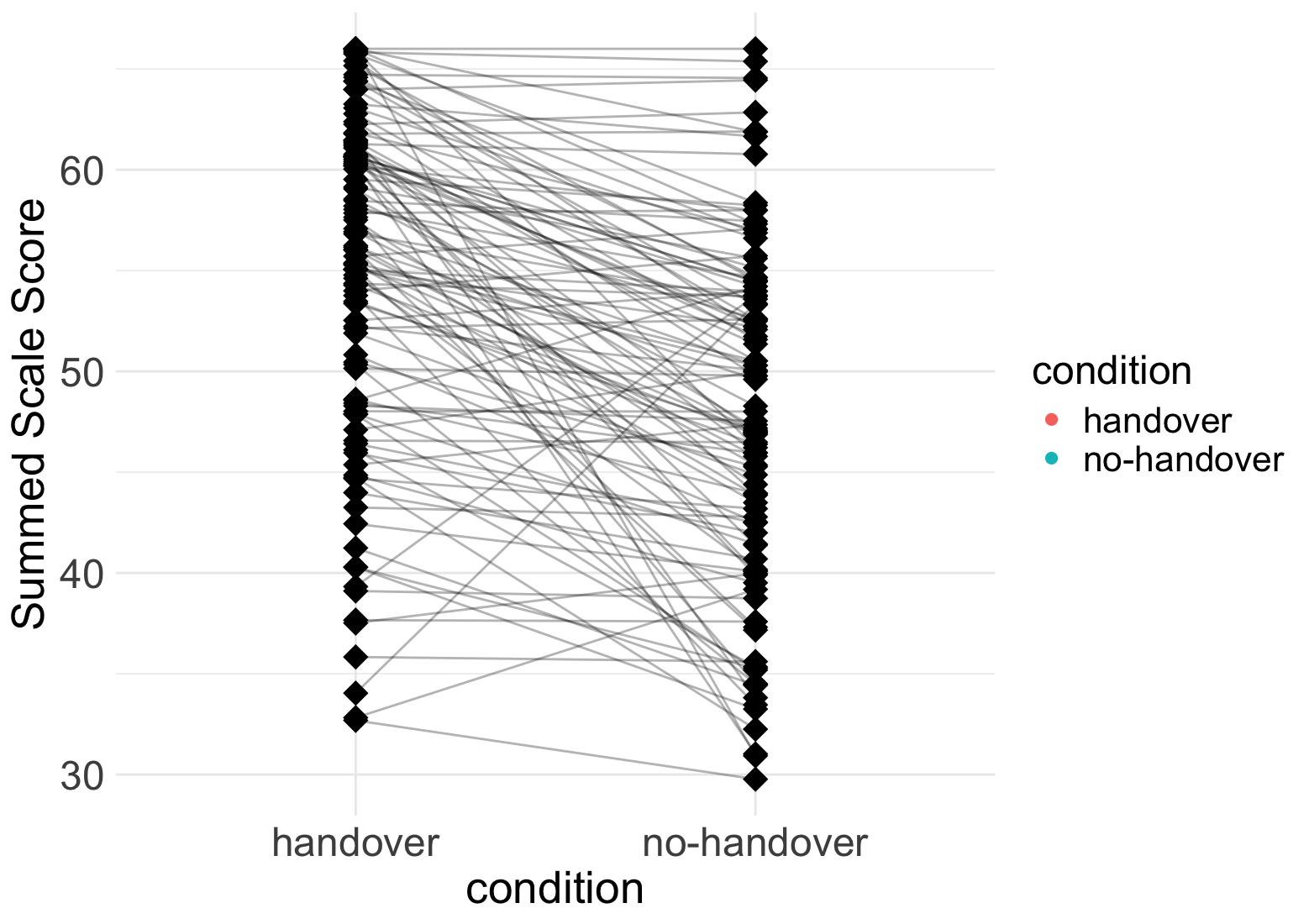}
    \caption{Exploratory comparison of handover and no-handover for personCharacteristics = aged.}
    \label{handover_paired_plot}
    %\vspace{-0.4cm}
\end{figure}

A paired t-test was performed to examine the mean differences in the Summed Scale Score between the handover and no-handover conditions. Participants rated the handover condition more positive (M = 54.67, SD = 8.42) than the no-handover condition (M = 48.04 SD = 8.61), t(111) = 8.81, p $<$ .001. Cohen's effect size (d = 0.83) indicates a large effect, thus we suggest that future research on the assessment of handover compared to no-handover conditions especially for elderly collaboration partners of robots is interesting. The results are illustrated by a box plot (\autoref{handover_boxplot}) and a paired plot (\autoref{handover_paired_plot}), which show the distribution of ratings in both conditions as well as the individual changes per participant.

\section{Discussion}\label{sec12}

We examined how participants assess a human-robot collaboration from the observer's perspective with regard to the combination of robot behavior and human physical characteristics. We investigated furthermore whether prior general reflection on human-robot collaboration using cognitive-affective mapping (CAM) leads to differences in the assessment of the specific variants of human-robot collaboration. 

To answer RQ-HRC and test the corresponding hypotheses (H2 to H9), we aggregated the data from the CAM and control groups. In the following, we discuss the differences we observed in the assessments of the variants of the human-robot collaboration. 

Antisocial robot behavior was rated significantly lower than all other behaviors, supporting hypothesis H2. This can be attributed to the expectation that, regardless of the human partner, robots should engage in a manner that actively involves humans in the interaction, especially when collaboration is anticipated. Effective human-robot collaboration is often assessed through an efficiency lens, meaning that humans expect robots to facilitate task completion faster and more effectively. From this perspective, antisocial behavior may be perceived as counterproductive, potentially hindering workflow and reducing overall task efficiency. 

Regarding the results of the hypothesis H3, it is possible to assume that participants believe that midFluency represents a smoother and more natural collaboration for young humans, as the robot provided enough time for the humans to complete the task without exerting pressure on them. This suggests that participants may prefer a balance between robotic assistance and human agency, rather than rigid alternation or highly autonomous robot behavior. Additionally, the fact that midFluency was rated more positively than alternating, even when compared to maxFluency, can indicate that excessive fluency might be perceived as too dominant or intrusive. 

For hypotheses H4, H5, H7, and H8, we found no significant differences. This can be due to the fact that the variations between the interactions presented in those conditions were not particularly noticeable or impactful. As a result, the participants may not have perceived the differences strongly enough to influence their judgments. 

The results provided partial support for H6. These findings suggest that, in general, observers preferred collaborations in the handover condition rather than the no-handover condition. This may be because the handover condition represents a more natural form of collaboration, allowing for more direct and intuitive interaction between humans and the robot. By engaging in physical exchange, participants may have felt a greater sense of involvement and control in the task, reinforcing the perception of effective teamwork. However, this result does not align with midFluency in the handover condition. This may be due to the interaction being influenced by the difference in the robot behavior resulting from the sorting scheme. Given that human-robot collaboration occurs in a direct way, the robot's speed in completing the task might not match the pace at which the human participant is accustomed to working. This could have lead to the observing study participants perceiving a mismatch between the efficiency and fluency of the collaboration.

Based on the results, we were also able to partially accept hypothesis H9. On the one hand, the difference found for the disabled and elderly human collaborator in the maxFluency condition in both the handover and no-handover conditions can be attributed to the type of person characteristic analyzed in the study. The disabled male was represented by a person walking with crutches, and since the task primarily involved hand usage, this kind of person characteristic might not face significant challenges when interacting with the robot under the maxFluency condition. Indeed, the videos of the disabled male indicated similar timing as young female and young male collaborators, and did not show any major obstacles in their collaboration with the robot.

On the other hand, the negative ratings of the maxFluency condition in both the handover and no-handover conditions for the collaborations with the aged female may be influenced by the fact that, compared to all other conditions, the maxFluency condition demands the highest levels of efficiency and speed for the human collaborator. This could result in increased stress and pressure for the human partners. In the observing participants, the demand for quick decisions and actions in such a condition could therefore evoke feelings of discomfort, leading to a more negative perception of the interaction. From the perspective of the observing participants, it seems that a direct handover cannot outweigh the disadvantages of the maxFluency condition for older people.

Overall, the participants rated the variations of the human-robot collaboration positively, with a mean sum score of 50.84 (with a possible maximum score of 66.00).\footnote{To what extent our scenario text influenced the positive ratings was beyond the scope of our assessment.}

The fact that some of our hypotheses, specifically those related to acceptability expectations, were not supported by the data highlights the inherent complexity of collecting social data to inform the development of assistive robots. It underscores the challenges of designing human-robot collaborations that meaningfully reflect the diverse and context-dependent nature of interaction types between humans and machines. This outcome underscores how challenging it remains, even for interdisciplinary teams that integrate diverse perspectives, to develop methods that adequately reflect the preferences, expectations, and contextual needs of different user groups. Notably, during the planning phase of this study, collaboration with computer scientists played a key role in the development of realistic scenarios aimed at simulating more complex and natural forms of human-robot collaboration that could be trained with social data.  However, despite these efforts, our findings suggest that aligning technical design with user-centered values in socially embedded contexts remains a critical and ongoing challenge in the field of social robotics.

While the data did not confirm some of our initial hypotheses, the analysis nonetheless yielded valuable insights into how individuals' preferences and perceptions can inform the development of robotic systems that are more adaptable, ethical, and responsive to users’ needs. Conducting this type of research requires in-depth analysis to uncover meaningful differences in robot behavior that are not always immediately apparent in everyday social contexts. A closer examination revealed subtle yet significant variations that can meaningfully guide the design of context-sensitive strategies for human-robot collaboration.

Therefore, these nuanced findings underscore the challenges and importance of adopting a human-centered approach, one that captures behavioral specificities of robots that might otherwise go unnoticed by users in typical social settings. By identifying these less visible changes in robot behavior, researchers can develop more inclusive, responsive, and ethically grounded robotic systems that better reflect the realities and expectations of users.
 
Regarding RQ-CAM, our results showed that drawing a CAM did not significantly change participants' perceptions of the specific human-robot collaboration in general. However, the results of the exploratory analysis suggest that creating a cognitive-affective map prior to the assessment of the human-robot collaboration influences the perception depending on the combination of human condition and robot behavior, as well as some of the queried subscales. Affective and social subscales, such as interpersonal fairness and likability, were particularly affected. This suggests that CAM sensitizes participants with respect to social and interpersonal aspects of interaction, which can lead to more critical or differentiated judgments. Functional subscales such as System Performance or Perceived Usefulness were not influenced .

%The differences in the ratings occurred most frequently in conditions with aged human collaboration partners (4 out of 7) and young humans (4 out of 14), while they were only significant in one out of seven conditions with disabled humans. This possibly indicates that reflection using CAM can lead to a change in perception, particularly when the interaction with aged humans is presented. Here, the participants seem to be more sensitized to social justice and fairness through the CAM—partly with more negative, partly with more positive ratings, depending on the robot behavior.
%For young humans, midFluency and alternating items in particular lead to more negative ratings after CAM, while positive effects can also occur in the maxFluency handover condition. In the case of disabled humans, it is noticeable that a significantly more negative rating only occurs in a specific handover condition—possibly because the discrepancy between the needs of the person depicted and the behavior of the robot is particularly clear here.

The present exploratory analyses indicate that prior reflection on cognitive and affective aspects of human-robot collaboration through CAM might be linked to condition-specific differences in the assessment of human-robot collaboration tasks. The exploratory results suggest that for vulnerable groups such as older adults or people with disabilities, sensitization through CAM could potentially foster a more nuanced perception of fairness and appropriateness of robot use. These initial observations may provide valuable starting points for future confirmatory research and for informing the design of user studies and social robots adapted to different user groups.

%The results propose that prior reflection of cognitive and affective aspects of human-robot collaboration through CAM can influence the assessment of human-robot collaboration tasks in a condition-specific way. Especially for vulnerable groups such as the aged or disabled humans, sensitization through CAM seems to promote a more differentiated perception of fairness and appropriateness of robot use. This has implications for the design of user studies and the development of social robots tailored to different user groups. Our findings are encouraging in two respects. First, they illustrate that it may be worthwhile to apply reflexive methods in human-robot collaboration studies in the context of a value-sensitive and human-centered approach to technology development to inform and enrich user perceptions. Second, they suggest that cognitive-affective maps could be suitable for this purpose. To confirm these assumptions, further research is needed.

\section{Conclusion}\label{sec13}

This study offers valuable insights into how human characteristics influence the perception and acceptance of collaborative robots on a specific scenario. By analyzing participant´s assessment of human-robot collaboration across varying combinations of robot behavior and human physical conditions, we identified certain preferences for certain HRC conditions. Our main findings revealed that antisocial robot behavior was rated significantly more negatively than all other behaviors, that the protected personal characteristic ‘aged’ in the human collaboration partner predicted that participants were more sensitive in their ratings, and that the handover conditions were rated significantly more positively than the no-handover conditions. These findings suggest that humans may have distinct preferences when collaborating with robots, which can be influenced by the context of the interaction and the robot's behavior. 

With this work, we aim to contribute to the field of social robotics from both a human-centered design method and an interdisciplinary perspective. By examining how different factors shape human-robot collaboration, we hope to offer valuable insights into designing more effective, ethical, user-friendly, and value-sensitive robotic systems.

\backmatter

\bmhead{Supplementary information}

In the supplementary material, we present additional results of an exploratory analysis regarding RQ-CAM, in which we examined in detail the conditions of human-robot collaboration that produced significant differences in specific subscales when comparing the CAM and control groups.

\bmhead{Funding}

This work was funded by the Carl Zeiss Foundation with the ReScaLe project, the German Research Foundation (DFG) Emmy Noether Program grant number 468878300, and the BrainLinks-BrainTools center of the University of Freiburg.

\bmhead{Data Availability}
The data that support the findings of this study are available from the University of Freiburg, but restrictions apply to the availability of these data, which were used under the consent signed by the participants and approved by the University of Freiburg Research Ethics Board to protect participants’ privacy. The data could be made available from the authors upon reasonable request and with the permission of the University of Freiburg Research Ethics Board.

\section*{Declarations}

\bmhead{Ethics Approval and Consent to Participate}

Before the experiment, participants received all the information and gave informed consent to participate in the study.

\bmhead{Conflict of Interest}

The authors declare to have no financial or non-financial conflict of interest.

{
\footnotesize
\bibliography{sn-bibliography}}% common bib file

\section{Supplementary Material}

In this section, we present the results of an exploratory analysis regarding RQ-CAM, in which we examined in detail the conditions of human-robot collaboration that produced significant differences in specific subscales when comparing the CAM and control groups.

\begin{table*}
    \centering
    \caption{Significant results of the comparison between CAM and control group.}
    \label{tab:independentSamplesTTest}
    \footnotesize
    \setlength{\tabcolsep}{2pt} % noch engerer Spaltenabstand
    \renewcommand{\arraystretch}{0.9} % etwas engerer Zeilenabstand
    \begin{tabular}{p{3.5cm} l l c c c p{1cm} p{1cm} p{1cm} p{1cm}}
        \toprule
        \textbf{Scale} & \textbf{Condition} & \textbf{Test} & \textbf{Statistic} & \textbf{df} & \textbf{p} & \textbf{Effect Size} & \textbf{SE Effect Size} & \multicolumn{2}{p{2cm}}{\textbf{95\% CI for Effect Size}} \\
        \cmidrule(lr){9-10}
        & & & & & & & & \textbf{Lower} & \textbf{Upper} \\
        \midrule
        interpersonal fairness & FYI & Student & -2.250 & 26.000 & 0.033 & -0.851 & 0.411 & -1.619 & -0.067 \\
         &  & Welch & -2.250 & 25.719 & 0.033 & -0.851 & 0.411 & -1.619 & -0.067 \\       
         &  & Mann-Whitney & 50.500 & & 0.030 & -0.485 & 0.219 & -0.746 & -0.094 \\
         interpersonal fairness & FYMH & Student & -2.211 & 26.000 & 0.036 & -0.836 & 0.410 & -1.603 & -0.054 \\
          &  & Welch & -2.211 & 18.841 & 0.040 & -0.836 & 0.410 & -1.613 & -0.039 \\
          &  & Mann-Whitney & 60.000 &  & 0.082 & -0.388 & 0.219 & -0.688 & 0.026 \\     
        interpersonal fairness & MYM & Student & -2.317 & 25.000 & 0.029 & -0.892 & 0.423 & -1.678 & -0.090 \\
         &  & Welch & -2.272 & 18.886 & 0.035 & -0.883 & 0.422 & -1.678 & -0.068 \\
         &  & Mann-Whitney & 48.500 &  & 0.040 & -0.467 & 0.223 & -0.740 & -0.062 \\
        perceived usefulness & MYM & Student & -2.362 & 25.000 & 0.026 & -0.910 & 0.424 & -1.697 & -0.106 \\
         &  & Welch & -2.336 & 22.040 & 0.029 & -0.905 & 0.424 & -1.696 & -0.095 \\
         &  & Mann-Whitney & 46.000 &  & 0.029 & -0.495 & 0.223 & -0.756 & -0.098 \\
        godspeed III: likabilty & MYM & Student & -2.805 & 25.000 & 0.010 & -1.080 & 0.440 & -1.883 & -0.260 \\
         &  & Welch & -2.721 & 14.541 & 0.016 & -1.063 & 0.438 & -1.895 & -0.202 \\
         &  & Mann-Whitney & 55.000 &  & 0.073 & -0.396 & 0.223 & -0.697 & 0.025 \\
         system performance & MYFH & Student & 2.722 & 26.000 & 0.011 & 1.029 & 0.425 & 0.229 & 1.811 \\
         &  & Welch & 2.722 & 17.914 & 0.014 & 1.029 & 0.425 & 0.204 & 1.830 \\
         &  & Mann-Whitney & 145.500 &  & 0.022 & 0.485 & 0.219 & 0.094 & 0.746 \\
        attitude toward using the robot & MYFH & Student & 2.517 & 26.000 & 0.018 & 0.951 & 0.419 & 0.159 & 1.727 \\
         &  & Welch & 2.517 & 18.251 & 0.021 & 0.951 & 0.419 & 0.139 & 1.742 \\
         &  & Mann-Whitney & 140.500 &  & 0.051 & 0.434 & 0.219 & 0.029 & 0.716 \\
        general rating & MYFH & Student & 2.518 & 26.000 & 0.018 & 0.952 & 0.419 & 0.159 & 1.728 \\
         &  & Welch & 2.518 & 18.625 & 0.021 & 0.952 & 0.419 & 0.140 & 1.741 \\
         &  & Mann-Whitney & 140.000 &  & 0.043 & 0.429 & 0.219 & 0.023 & 0.713 \\
         interpersonal fairness & FAA & Student & 2.129 & 26.000 & 0.043 & 0.805 & 0.407 & 0.025 & 1.570 \\
         &  & Welch & 2.129 & 25.354 & 0.043 & 0.805 & 0.407 & 0.024 & 1.570 \\
         &  & Mann-Whitney & 140.500 &  & 0.053 & 0.434 & 0.219 & 0.029 & 0.716 \\
         godspeed III: likabilty & FAF & Student & -2.243 & 25.000 & 0.034 & -0.864 & 0.421 & -1.648 & -0.065 \\
         &  & Welch & -2.239 & 24.620 & 0.034 & -0.863 & 0.421 & -1.647 & -0.063 \\
         &  & Mann-Whitney & 47.000 &  & 0.033 & -0.484 & 0.223 & -0.749 & -0.084 \\
         general rating & FAFH & Student & 2.095 & 26.000 & 0.046 & 0.792 & 0.407 & 0.014 & 1.556 \\
         &  & Welch & 2.095 & 19.771 & 0.049 & 0.792 & 0.407 & 0.002 & 1.563 \\
         &  & Mann-Whitney & 135.500 &  & 0.086 & 0.383 & 0.219 & -0.032 & 0.685 \\
         perceived usefulness & FAIH & Student & -2.287 & 25.000 & 0.031 & -0.881 & 0.422 & -1.666 & -0.080 \\
         &  & Welch & -2.264 & 22.425 & 0.034 & -0.876 & 0.422 & -1.664 & -0.071 \\
         &  & Mann-Whitney & 55.500 &  & 0.081 & -0.390 & 0.223 & -0.694 & 0.032 \\
        quality of interaction & FAIH & Student & -3.255 & 25.000 & 0.003 & -1.254 & 0.457 & -2.074 & -0.413 \\
         &  & Welch & -3.231 & 23.320 & 0.004 & -1.249 & 0.456 & -2.073 & -0.403 \\
         &  & Mann-Whitney & 31.500 &  & 0.004 & -0.654 & 0.223 & -0.841 & -0.326 \\
         interpersonal fairness & MDFH & Student & -2.787 & 25.000 & 0.010 & -1.073 & 0.439 & -1.875 & -0.253 \\
         &  & Welch & -2.703 & 14.585 & 0.017 & -1.056 & 0.437 & -1.887 & -0.197 \\
         &  & Mann-Whitney & 47.500 &  & 0.036 & -0.478 & 0.223 & -0.746 & -0.076 \\
        attitude toward using the robot & MDFH & Student & -2.150 & 25.000 & 0.041 & -0.828 & 0.418 & -1.609 & -0.032 \\
         &  & Welch & -2.100 & 17.159 & 0.051 & -0.818 & 0.417 & -1.610 & -0.005 \\
         &  & Mann-Whitney & 55.500 &  & 0.088 & -0.390 & 0.223 & -0.694 & 0.032 \\
        godspeed III: likabilty & MDFH & Student & -2.079 & 25.000 & 0.048 & -0.801 & 0.416 & -1.580 & -0.007 \\
         &  & Welch & -2.030 & 17.111 & 0.058 & -0.791 & 0.415 & -1.580 & 0.019 \\
         &  & Mann-Whitney & 61.000 &  & 0.146 & -0.330 & 0.223 & -0.656 & 0.101 \\

        \bottomrule
    \end{tabular}
\end{table*}

\subsection{CAM-related Exploratory Analysis}

An exploratory analysis of the data revealed some significant and marginally significant differences, with medium to high effect sizes, on individual scales between the CAM and control groups, depending on single comparisons of the HRC. \autoref{tab:independentSamplesTTest} gives an overview. Among the 14 conditions examined with young humans (female/male), 4 showed significant differences between the CAM and control groups. Notably, for the young men in the midFluency (MYM) condition in particular, the CAM group rated Interpersonal Fairness, Perceived Usefulness, and Likability of the robot significantly lower than the control group (e.g., Interpersonal Fairness: d = -0.892, p = 0.029). Young women in the alternating items (FYI) and midFluency handover (FYMH) conditions were also found to have lower values for Interpersonal Fairness, although some results were marginal.
An exception is the maxFluency handover (MYFH) condition for young men, in which the CAM group gave significantly higher ratings for System Performance, Attitude towards Use, and the general rating (e.g., System Performance: d = 1.029, p = 0.011).

Of 7 conditions with aged humans, 4 showed significant differences. In the antisocial condition (FAA), the CAM group rated Interpersonal Fairness significantly higher (d = 0.805, p = 0.043). In alternating items (FAF), the rating of Likability was significantly lower (d = -0.864, p = 0.034). In the handover condition maxFluency (FAFH), the general rating by the CAM group was significantly more positive (d = 0.792, p = 0.046). In the alternating items handover (FAIH) condition, the CAM group rated Perceived Usefulness and Quality of Interaction significantly lower (e.g., Quality of Interaction: d = -1.254, p = 0.003).

Among the seven conditions involving individuals with disabilities, only one condition (maxFluency handover, MDFH) showed significantly lower ratings from the CAM group on several subscales: Interpersonal Fairness, Attitude towards Use, and Likability (e.g., Interpersonal Fairness: d = -1.073, p = 0.010). The most frequently affected scales were Interpersonal Fairness and Likability (Godspeed III), followed by Perceived Usefulness, System Performance, Quality of Interaction, Attitude towards Use, and General Rating. The direction of the effects varies: While in most cases the ratings are more negative in the CAM group, there are also conditions with significantly more positive ratings by the CAM group.

\end{document}